\documentclass{article}
\usepackage{log_2023}  
\usepackage[numbers,compress,sort]{natbib}

\usepackage{amsmath,amsfonts,bm}

\def\eqref#1{equation~\ref{#1}}

\def\1{\bm{1}}

\def\vh{{\bm{h}}}

\def\vx{{\bm{x}}}

\DeclareMathAlphabet{\mathsfit}{\encodingdefault}{\sfdefault}{m}{sl}
\SetMathAlphabet{\mathsfit}{bold}{\encodingdefault}{\sfdefault}{bx}{n}

\def\gB{{\mathcal{B}}}

\def\gT{{\mathcal{T}}}

\usepackage[utf8]{inputenc} 
\usepackage[T1]{fontenc}    
\usepackage{url}            
\usepackage{booktabs}       
\usepackage{amsfonts}       
\usepackage{nicefrac}       
\usepackage{microtype}      
\usepackage{xcolor}         
\usepackage{amsmath}
\usepackage{multirow}
\usepackage{subcaption}
\usepackage{graphicx}
\usepackage{duckuments}          

\usepackage{booktabs}
\usepackage{array}

\title[Neural Algorithmic Reasoning for Combinatorial Optimisation]{Neural Algorithmic Reasoning for Combinatorial Optimisation}

\author[D. Georgiev et al.]{%
  \hspace{30pt}Dobrik Georgiev\thanks{Equal contribution -- order is alphabetical.}\\
  \hspace{30pt}University of Cambridge\\
  \hspace{30pt}\texttt{dgg30@cam.ac.uk}\\
  \And
  Danilo Numeroso$^*$ \\
  Università di Pisa \\
  \texttt{danilo.numeroso@phd.unipi.it} \\
  \And
  Davide Bacciu \\
  Università di Pisa \\
  \texttt{davide.bacciu@unipi.it} \\
  \And
  Pietro Liò \\
  University of Cambridge \\
  \texttt{pl219@cam.ac.uk} \\
}

\graphicspath{{./figures/}}

\begin{document}

\maketitle

\begin{abstract}
    Solving NP-hard/complete combinatorial problems with neural networks is
    a challenging research area that aims to surpass classical approximate
    algorithms. The long-term objective is to outperform hand-designed heuristics
    for NP-hard/complete problems by learning to generate superior solutions solely
    from training data. Current neural-based methods for solving CO problems often
    overlook the inherent ``algorithmic'' nature of the problems. In contrast,
    heuristics designed for CO problems, e.g. TSP, frequently leverage
    well-established algorithms, such as those for finding the minimum spanning
    tree. In this paper, we propose leveraging recent advancements in neural
    algorithmic reasoning to improve the learning of CO problems. Specifically, we
    suggest pre-training our neural model on relevant algorithms before training it
    on CO instances. Our results demonstrate that by using this learning setup, we
    achieve superior performance compared to non-algorithmically informed deep
    learning models.
\end{abstract}

\section{Introduction}


\footnotetext[0]{The source code can be found at \url{https://github.com/danilonumeroso/conar}.}

Combinatorial problems have immense practical applications and are the backbone
of modern industries. For example, the Travelling Salesman Problem (TSP) has
applications as diverse as logistics \citet{baniasadi2020transformation}, route
optimisation \citep{xu2019brief}, scheduling to genomics
\citep{sankoff1998multiple} and systems biology \cite{johnson2006protein}. The
problem has been studied by both theoretical computer scientists
\cite{balas1983branch, karlin2022slightly} and the ML community 
\cite{vinyals2015, kool2019attention, joshi2022learning} from different perspectives.
The former targets the development of hand-engineered approximate
algorithms, which trade-off optimality for computational
complexity, whereas the latter's ultimate objective is to learn such
algorithms instead of manually designing them, i.e. Neural Combinatorial Optimisation (NCO). Hence, learning to solve such a problem and
pushing the performance beyond that of hand-crafted heuristics, such as the
Christofides' \citep{christofides1976worst} algorithm, is a fascinating goal
that can have a direct impact on the aforementioned industrial fields. 

Existing works attempt to learn solutions for CO problems by either supervised
learning \cite{joshi2022learning} or reinforcement learning \cite{kool2019attention}, exploiting standard deep learning methods. Although
many combinatorial problems have natural ``algorithmic'' solutions \cite{christofides1976worst, bhattacharya2017fully}, leveraging algorithmic knowledge has never been explored in prior work.

For example, the Christofides' algorithm follows several steps to construct a solution for the TSP. First, it computes the minimum spanning tree (MST) and subsequently applies a matching algorithm to it. In light of this observation, we argue that knowledge of algorithms may be useful also when addressing combinatorial problems through the use of neural networks. Our hypothesis is that a neural network that possesses knowledge of, say, the minimum spanning tree, might be able to generalise better when learning to solve the TSP. In this context, recent advancements in Neural Algorithmic Reasoning (NAR) \cite{velickovic2021neural} neural networks have been shown to effectively learn and reproduce the execution of classical algorithms \cite{velickovic2021neural}, such as the Prim's algorithm \cite{prim1957shortest}. Moreover, current ``neural algorithmic reasoners'' can learn multiple algorithms at once, showing positive transfer knowledge across them. This yields to better results compared to learning a single algorithm in isolation. 

Motivated by this, in this paper, we explore the applicability of NAR in the
context of NCO. So far, most of the prior work in neural
algorithmic reasoning focuses on learning algorithms and solving problems
within the class of complexity of problems solvable in polynomial time, i.e. P.
In our settings, we aim to solve challenging combinatorial problems, e.g., TSP,
which are known to be NP-complete/hard. Hence, we try to transfer algorithmic
knowledge from algorithms and problems that are in P to NP-complete/hard
problems. To the best of our knowledge, we are the first to explore
transferring from algorithmic reasoning to NP-complete/hard combinatorial
optimisation problems.

In general, the contributions of our work can be summarised as follows:
(i) we show that \textbf{algorithms are a good inductive bias in the context of NCO} and leads to good OOD generalisation w.r.t. algorithm-agnostic baselines;
(ii) we investigate and \textbf{study how different NAR transfer learning settings perform} when applied on the P$\rightarrow$NP scenario. Interestingly, the best setting is pre-train + finetune -- a method that was previously thought inferior in NAR;
(iii) we \textbf{highlight shortcomings} of current neural algorithmic reasoners, opening up new potential areas for research.

\section{Related work}
Neural Combinatorial Optimisation (NCO) is a long-standing research
area that aims to exploit the predictive power of deep neural networks
to help solve CO problems. Pioneering work from Hopfield and Tank
\cite{hopfield1985} dates back to the 1980s, where the authors
designed a specialised loss function to train a rudimentary neural
network to solve instances of TSP. Building from there, practitioners
have come up with increasingly complex models to solve CO problems,
starting from Self-Organising Maps (SOMs) until modern deep learning
architectures. The Pointer Networks (Ptr-Nets) \cite{vinyals2015}
represent the first attempt to solve a CO task, namely the TSP, with a
pure deep learning approach. Ptr-Nets are sequence-to-sequence
recurrent models that process a sequence of 2D geometrical points,
e.g. coordinates of TSPs, and ``translate'' them to a sequence of
nodes to be visited, i.e. a {\it tour} $\pi$. The original model was
trained in a supervised way, even though some different training
setups have been explored by later works, such as actor-critic
reinforcement learning \cite{bello2017} and REINFORCE
\cite{deudon2018}. However, the community quickly realised the benefit
of using structure-aware learning models, such as Graph Neural
Networks (GNNs). Many CO problems, in fact, have natural
representations as graphs rather than sequences of points in
geometrical spaces (see \autoref{sec:background}, TSP paragraph). In
particular, Kool et al. \cite{kool2019attention} and Joshi et
al. \cite{joshi2022learning} developed a similar encoder-decoder to
Ptr-Nets but treat the input as a graph, incorporating Graph Attention
Network (GAT) \cite{velickovic2018GAT} style updates in their respective
architectures. However, the entirety of the proposed approaches in NCO
treat the problem as a ``standard'' predictive task, without taking
into account that algorithms play an important role in classical
computer science heuristics, e.g. Christofides' algorithm for
TSPs. Differently from all prior works, we incorporate algorithmic
reasoning knowledge in our learnt predictor, i.e. a Message-Passing
Neural Network (MPNN), by leveraging neural algorithmic reasoning
\cite{velickovic2021neural}.

As introduced in \autoref{sec:background}, Neural Algorithmic
Reasoning (NAR) involves developing neural models and learning setups
to facilitate the encoding of algorithms directly in models'
weights. Starting from some early work
\cite{velickovic2020execution} that aimed to demonstrate the
applicability of GNNs to reproduce classical algorithm steps, the
community has then applied these concepts to a variety of
domains. Deac et al. \cite{deac2021neural} successfully
trained an algorithmic reasoner to planning problems, by reproducing
steps of Bellman optimality equation in a reinforcement learning
setting. Xhonneux et al. \cite{xhonneux2021transfer} studied
different ways of transferring algorithmic knowledge between
algorithms. In particular, they train algorithmic reasoners on a set
of algorithms and study the effect of: (i) training on all algorithms
simultaneously ({\it multi-task learning}); (ii) pre-train on a subset
of algorithms and {\it finetune} on others; (iii) freeze some
parameters of the pre-trained network and learn the remaining; (iv)
utilise multiple processors, one of which is frozen. In our work, we
experimented with transferring algorithmic knowledge by using and extending
these setups. \citet{numeroso2023DAR} train an algorithmic
reasoner to solve MaxFlow/MinCut by exploiting the duality of the two
optimisation problems. However, the combinatorial problem therein is
in P. Differently from them, we try to generalise and transfer
algorithmic knowledge from problems that are in P, i.e. shortest path
and minimum spanning tree, to harder problems such as the TSP, which
is known to be NP-{\it hard}.

\section{Background} \label{sec:background}
\paragraph{Graph Neural Networks}
Graph Neural Networks \cite{bacciu2020introduction}, are a type
of deep neural networks designed to operate within a structured input
domain, i.e. graphs, by relying on a local context diffusion
mechanism. Specifically, let $G=(V,E)$ be a graph with $V$ and $E$
being respectively the set of nodes and edges. Then, GNNs process
graphs by propagating information at a node-level, following a
message-passing algorithm as follows:
\begin{equation}
    \vh^{(\ell+1)}_v = \phi^{(\ell+1)} \big (
        \vh^{(\ell)}_v,
        \Psi(\{ \psi^{(\ell+1)}(\vh^{(\ell)}_u) \mid u \in \mathcal{N}(v)\})
    \big )
    \label{eq:node-aggr}
\end{equation}
where $\phi$ and $\psi$ are parameterised transformations,
$\vh^{(\ell)}_v$ is the node representation at layer/iteration $\ell$,
with $\vh^{(0)}_v=\vx_v$ where $\vx_v$ is a vector of initial
features. Function $\Psi$ aggregates the information flowing to node $v$ from its neighbourhood $\mathcal{N}(v)$, and is chosen to be a
permutation-invariant function, i.e. a different ordering does not
change the final output.

\paragraph{Neural execution of algorithms} Neural Algorithmic Reasoning (NAR)
\cite{velickovic2021neural} is a developing research area that seeks to build
algorithmically-inspired neural networks. In classical algorithms such as
Bellman-Ford \citep{richard1958routing, ford1956network} and Prim
\citep{prim1957shortest}, solutions are typically constructed through a series
of iterations. When executing algorithm $A$ on input $\vx$, it produces an
output $A(\vx)$ and defines a \emph{trajectory} of intermediate steps
$A_t(\vx)$. Unlike traditional machine learning tasks that focus solely on
learning an input-output mapping function $f: X \rightarrow A(X)$, NAR aims to
impose constraints on the hidden dynamics of neural models to replicate
algorithmic behaviours. This is commonly achieved by incorporating supervision
on ${A_t(\vx)}$ \citep{velickovic2020execution} in addition to any supervision
on $A(\vx)$.

The encode-process-decode architecture \citep{hamrick2018relational} is
considered the preferred choice when implementing NAR models
\cite{velickovic2021neural}. The architecture is a composition
of three learnable components $net = g \circ p \circ f$, where $f$ and $g$ are
encoding and decoding functions (usually as simple as linear transformations),
and $p$ is a suitable neural architecture called {\it processor} that mimics
the execution rollout of $A$. Typically, we choose $p$ to be the only part of
the network having non-linearities to ensure the majority of the architecture's
expressiveness lies within the processor. Furthermore, as shown in
\citet{ibarz2022generalist}, we are also able to learn {\it multiple}
algorithms (both graph and non-graph) at once, by having separate $f_A$ and
$g_A$ per an algorithm $A$, and sharing the processor $p$ across all
algorithms.

A fundamental objective of NAR is to achieve robust out-of-distribution (OOD)
generalisation. Typically, models are trained and validated on ``small'' graphs
(e.g., up to 20 nodes) and tested on larger graphs. This is inspired by classical
algorithms' size-invariance, where correctness of the solution is maintained
irrespective of the input size. The rationale behind testing on larger graphs
is to assess whether the reasoner has truly captured the algorithm's behaviour
and can extrapolate well beyond the training data, and past the easily
exploitable shortcuts in it. Evaluating models on OOD data reveals their
ability to generalize and provide accurate solutions in diverse scenarios.


\paragraph{CLRS-30} 

The \emph{CLRS-30 benchmark} \citep{velivckovic2022clrs}, or CLRS-30, comprises
30 iconic algorithms from the \textit{Introduction to Algorithms} textbook
\citep{CLRS}. The benchmark covers diverse algorithm types, including string
algorithms, searching, dynamic programming, and graph algorithms. \emph{All}
data instances in the CLRS-30 benchmark are represented as graphs and are
annotated with \emph{input}, \emph{output} and \emph{hint} features and an
associated position.  Denote the dimensionality of a feature as $F$ and $|V|$
as $N$. Input/output node features have the shape $N\times F$, edge features --
$N\times N \times F$, graph features -- only $F$. Hints encapsulate time series
data of algorithm states. Like inputs/outputs, they include a temporal
dimension which also indicates the duration of execution. All features fall
into 5 types: \texttt{scalar}, \texttt{categorical}, \texttt{mask} (0/1 value),
\texttt{mask\_one} (only a single position can be 1), and \texttt{pointer}.
Each feature type has an associated loss to be used when training the neural
network \citep[cf.][]{velivckovic2022clrs}.

\paragraph{Travelling Salesman Problem} A TSP instance is a complete
undirected weighted graph $G=(V, E, w)$. A valid solution for a TSP is a {\it
tour} $\pi = v_0v_1 \dots v_{|V|}$, wherein $\pi_{0:|V|}$ is
a \emph{permutation} of the $n$ nodes -- the node visiting order before
returning to the starting node. Among all correct solutions, we usually seek the
one minimising the total cost of the tour $c^* = \min_{\pi}
\sum^{|V|-1}_{i=0}{w_{\pi_i\pi_{i+1}}}$.

\paragraph{Vertex K-center} The vertex k-center problem (VKC)
\citep{hakimi1964optimum} is defined as follows: given an undirected weighted
graph $G=(V, E, w)$ and a positive integer $k$, find a subset $C\subset V$
where $\lvert C \rvert \leq k$ in order to minimize $o = \max_{v\in V}{w(v,
C)}$. In this context, $w(v, C)$ denotes the distance from vertex $v$ to its
nearest center in subset $C$. In the original formulation, the problem involves
a complete graph and a metric function $w$. To introduce some diversity
compared to TSP, we consider (not necessarily fully) connected graphs and
necessitate the network to compute the shortest distances between nodes.



\section{Neural Algorithmic Reasoning for Combinatorial Optimisation}

\subsection{Selection of relevant algorithms}
\label{sec:algorithm-selection}

In our algorithm selection for pre-training, we prioritize specific
subtask-solving algorithms within the target combinatorial optimization
problem, like Prim's algorithm for TSP or Find-Min for VKC. We also consider
algorithms that monitor properties (e.g., shortest path) or belong to related
classes (e.g., greedy algorithms). Exclusion of algorithms is also vital. We
omit algorithms that the base reasoner struggles with (e.g., Floyd-Warshall for
VKC, see \autoref{tab:vkc-results}) or are resource-intensive (cf.
\autoref{app:runtime}) due to their CLRS-30 representation.
The final choice is Bellman-Ford and MST-Prim for TSP and Bellman-Ford,
min finding, activity selection \citep{gavril1972algorithms} and task
scheduling \citep{lawler2001combinatorial} for VKC. \autoref{app:algosel}
provides deeper insights into our algorithm selection process.  

\subsection{Choice of GNN architecture}
\label{subsec:gnn-choice}

The rationale behind selecting our GNN architecture stemmed from the necessity
of being able to learn to perform algorithms, in particular shortest path,
MST-Prim and Graham Scan. Consequently, we intentionally avoided utilizing
architectures such as GCN/GAT \citep{kipf2017semisupervised, velickovic2018GAT}
which are known to perform poorly in algorithmic reasoning
\citep{velickovic2020execution} or Pointer Networks \citep{vinyals2015} which
do not incorporate graph structure information. A further reason to avoid standard graph neural TSP architectures is that they usually have fixed depth, while algorithmic reasoning architectures are recurrent.

Despite narrowing down our potential architectures, there have been several
impactful architectures in the last few years. Testing each and every possible
architecture requires computational resources outside of our reach. In order to choose an appropriate architecture, we focused on several key properties the network should possess: \emph{accuracy} in graph algorithms and particular
Bellman-Ford and MST-Prim, \emph{scalability} and \emph{suitability} for TSP. The first requirement has led us to exclude the recently proposed 2-WL architecture by \citet{mahdavi2023towards}.
Lastly, Pointer Graph Networks \citep{velickovic2020pointer} are unusable for our purpose due to the nature of TSP. Persistent Message Passing \citep{strathmann2021persistent} is also inappropriate as it targets fundamentally different type of problems.

Based on the above analysis our final choice for a neural algorithmic reasoner is an MPNN architecture with graph features and a gating mechanism as in
\citet{ibarz2022generalist}:
\begin{align*}
    \mathbf{z}_i^{(t)} = \mathbf{u}_i^{(t-1)} \Vert \mathbf{h}_i^{(t-1)} \qquad &
    \mathbf{m}_i^{(t)} = \max_{1\leq j \leq n} f_m\left(\mathbf{z}_i^{(t)}, \mathbf{z}_j^{(t)}, \mathbf{e}_{ij}^{(t)} \right) \qquad
    \mathbf{\hat{h}}_i^{(t)} = f_r\left(\mathbf{z}_i^{(t)}, \mathbf{m}_i^{(t)}\right) \\
    \mathbf{g}_i^{(t)} = f_g(\mathbf{z}_i^{(t)}, \mathbf{m}_i^{(t)}) \qquad &
    \mathbf{h}_i^{(t)} = \mathbf{g}_i^{(t)} \odot \mathbf{\hat{h}}_i^{(t)} + (1-\mathbf{g}_i^{(t)}) \odot \mathbf{\hat{h}}_i^{(t)}
\end{align*}
where $\mathbf{u}_i^{(t)}$ are node input/hint features at timestep $t$,
$\vh_i^{(t)}$ is the latent state of node $i$ at timestep $t$ (with
$\vh_i^{(0)}=\mathbf{0}$), $\Vert$ denotes concatenation and $\odot$ denotes
elementwise multiplication. Further, $f_m$ is a message function, and $f_g$ and $f_r$ are gating-related functions of the model. Function $f_m$ is parametrised as an MLP, while $f_g$ and $f_r$ are linear projection layers. Inspired by the architecture of \citet{joshi2022learning} we also explored having edge hidden states. The state is updated similar to $\vh_i^{(t)}$ and is used in the message computation: $f_m\left(\mathbf{z}_i^{(t)}, \mathbf{z}_j^{(t)},
\mathbf{e}_{ij}^{(t)}, \vh_{ij}^{(t)} \right)$.




\subsection{Algorithmic knowledge integration}\label{subsec:transfer}

None of the algorithmic reasoning papers we surveyed experimented with NP-hard problems like TSP. Thus, we decided to test a variety of ways to expose our model to algorithmic knowledge, including those that have previously shown inferior performance \citep{xhonneux2021transfer}.  Denote a base task
$\mathcal{B}$, from which we want to transfer, and a target task $\mathcal{T}$,
which we want to infuse with the knowledge of $\mathcal{B}$. We explored the following integration strategies:
\begin{itemize}
    \item \textbf{Pre-train and freeze (PF)} -- this is the most common setting used
        when $\gB$ is an abstract algorithm and $\gT$ is a real-world task
        \citep{deac2020xlvin, numeroso2023DAR, velickovic2021neural}. It
        consists of pre-training a model on a concrete task (e.g. MST) and
        copying the \emph{processor} parameters into the real-world
        architecture's processor and freezing them so they cannot be changed.
    \item \textbf{Pre-train and fine-tune (PFT)} -- in the approach above the
        processor weights are frozen. We hypothesise this may not be suitable
        for our P to NP-hard transfer, so we provide results using the same
        transfer learning technique as above but allowing weights to be optimised.
    \item \textbf{2-processor transfer (2PROC)} \cite{xhonneux2021transfer} --
        this technique is a combination of the aforementioned two. When
        transferring from $\gB$ to $\gT$ we use 2 processors: one initialised
        with the pre-trained parameters and kept frozen and another randomly
        initialised and fine-tuned.
    \item \textbf{Multi-task learning (MTL)} -- this is a setting where the transfer
        from $\gB$ to $\gT$ is performed implicitly by training $\gB$ and $\gT$
        together \emph{but sharing the same processor}. This technique gave the best
        results in \citet{xhonneux2021transfer} and was also used when training
        with dual algorithms in \citet{numeroso2023DAR}.
\end{itemize}



\section{Evaluation}

\subsection{Data generation \& hyperparameter setup}

\paragraph{NAR} For our NAR training set we generated 10000
samples for each algorithm. For pre-training the NAR model for the TSP task we
generated each of our graphs by sampling points in 2D space (unit square) and
setting the edge weights of the graph equal to the Euclidean distance between
nodes. Eventually, graphs are all fully-connected. For VKC, we sample Erdős–Rényi \citep{erdos1960evolution} graphs with
$p=0.5$, excluding graphs that are disconnected. As it has been shown that
varying graph sizes leads to stronger reasoners
\citep{mahdavi2023towards,ibarz2022generalist}, the size of each graph was
uniformly chosen from $[8, 16]$. For generating the ground-truth targets and
trajectories, we employed the official code for the CLRS-30 benchmark. Although
we picked our reasoner based on the best validation score we also built a test set
in order to evaluate the capability of our reasoners to generalise. Both
validation and test sets comprehend 100 graphs per algorithm, but graph sizes
in the validation dataset are fixed at 16, while graphs in the test dataset are
of size fixed at 64.

\paragraph{TSP} To train our CO models, we adopted the
representation format of CLRS-30 (cf. \autoref{sec:background}). As input
features for TSP we picked \texttt{node mask} features representing the
starting node of the tour and \texttt{edge scalar} features for the distance
between nodes. Node coordinates were replaced with the distance matrix to make
our architecture invariant to rotations and translations. We defined our output
features as \texttt{node pointers}, where every node predicts a probability of
each other node being its predecessor in the TSP tour which is optimised using categorical cross entropy.  Lacking a TSP execution
trajectory, we used ``pseudo-hints'' from tracing the optimal tour for
experiments.  Regrettably, these pseudo-hint trials yielded poor results and
were excluded from our final setup.

For the TSP task we generated larger datasets of graphs of varying size. We
generated $100000$ training samples of each of the sizes $[10, 13, 16, 19, 20]$
($500 000$ total). While this seems much larger than the training data we have
in the P case, it is only half the amount of data of \citet{joshi2022learning}
and is several orders of magnitude less than \citet{kool2019attention}.
Further, graph sizes are much smaller than in previous works
\citep{joshi2022learning, khalil2017learning, kool2019attention} which may
sometimes reach or exceed $100$ nodes.

For validation we generated $100$ graphs of size $20$.  We further generated
$[1000,32,32,32,32,4]$ test graphs of sizes $[40,60,80,100,200,1000]$
respectively. Finally, we generated the optimal ground-truth tours using the
Concorde solver \cite{applegate2006concorde}.

\paragraph{VKC} VKC has a simpler ``specification'' -- input
\texttt{edge scalar} features for edge weights and output \texttt{node mask}
features signifying node inclusion in set $C$. 

For training/validation/testing, we kept $k=5$ constant and generated graphs of
the same sizes (except 1000) and numbers as for TSP. The main distinction from
TSP lies in our approach to obtaining the ``ground-truth'' solution: We observe
that the optimal value must be equivalent to the distance between two nodes
\citep{hochbaum1985best}. This insight enables us to conduct binary search and
leverage Gurobi \citep{gurobi} for solving the LP formulation of VKC's decision
variant\footnote{Given a cost $o$, is it possible to achieve a solution with
$k$ or fewer vertices that has a lower or equal cost}.

\paragraph{Hyperparameter setup} In all our experiments we instantiate our
models with latent dimensionality of 128. We train using batch size of 64 and
Adam \citep{kingma2015adam} optimiser with a learning rate of 0.0003, with no
weight decay.  We report standard deviations from our experiments for 5 seeds.
In transfer learning experiments, we use different algorithmic reasoners for
each seed. We train our algorithmic reasoners for 100 epochs, TSP models for
20, VKC models for 40.

\subsection{Solution Decoding} Neural networks may not always generate valid
outputs. For TSP, node pointers can lead to incomplete tours or those not
ending at the starting node. Therefore, for TSP, we employed beam search,
following the approach of \citet{joshi2022learning}, to identify the most
likely set of node pointers constituting a valid tour. Similarly, in VKC, the
neural network might not yield a valid solution (e.g., selecting more than $k$
vertices). To handle this, we post-processed the model's output by selecting
the top $k$ vertices based on the highest confidence (probability) of node
membership in set $C$.



\subsection{TSP}\label{subsec:TSP}

\paragraph{Benchmarks} To benchmark our model, we conducted comparisons against
various baselines. Firstly, we examined the performance of our model with and
without transfer learning, using the same hyperparameters. As additional
benchmarks, we compared our model to prior architectures proposed by
\citet{joshi2022learning,rampavsek2022recipe}, retraining them on the same
dataset utilized for our models. For \citet{joshi2022learning} we used the code
provided, setting the gating to true\footnote{\citet{joshi2022learning} mention
they use gating on p.5 of their paper}, and defaulting all others. The
choice of these previous models as benchmarks was motivated by their
comparable number of parameters to our own model. For \citet{rampavsek2022recipe} we used the official PyG \citep{Fey/Lenssen/2019} implementation provided with the same MPNN processor as ours as the \texttt{conv} parameter in order to be able to process edge features.

Furthermore, we included results from deterministic approaches for
comprehensive evaluation. These deterministic baselines consisted of a greedy
algorithm that always selected the lowest-cost edge to expand the tour, beam search with
a width of 1280, the
Christofides heuristic \citep{christofides1976worst} and the LKH heuristic \citep{helsgaun2000effective}. Additional benchmarking against constraint programming/mixed
integer programming is given in \autoref{app:gurobiowned}.

Finally, to evaluate the performance of all the tested models we use the
relative error w.r.t. the cost of the optimal tour $\pi^*$. Given
the optimal tour cost $c^*$ obtained by the Concorde solver, we compute the
relative error as $r=\frac{\tilde{c}}{c^*} -1$ where $\tilde{c}$ is the cost of
the tour $\tilde{\pi}$ predicted by the neural network.

\begin{figure}[t]
  \centering
  
  \begin{subfigure}{0.49\textwidth}
  \centering
    \includegraphics[width=\linewidth]{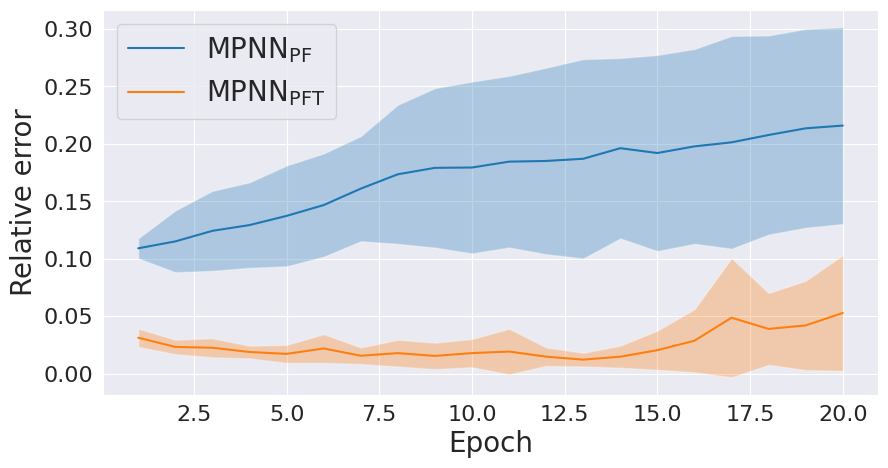}%
    \label{fig:subValPFTvsPF}%
    \caption{Validation relative error per epoch}%
  \end{subfigure}\hfill%
  \begin{subfigure}{0.49\textwidth}
  \centering
    \includegraphics[width=\linewidth]{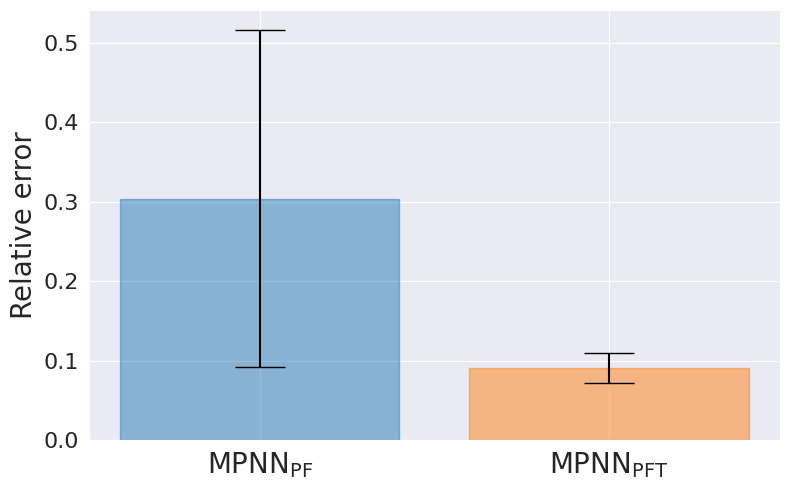}%
    \label{fig:subTestPFTvsPF}%
    \caption{Final test relative error}%
  \end{subfigure}%
  \caption{
      Standard transfer learning approaches are not applicable for our purposes:
      the resulting models are unable to generalise both in- and out-of-distribution. 
  }
  \label{fig:PFTvsPF}
\end{figure}

\begin{figure}[t]
  \centering 
  \begin{subfigure}[c]{0.49\textwidth}
    \centering
    \includegraphics[width=0.99\linewidth]{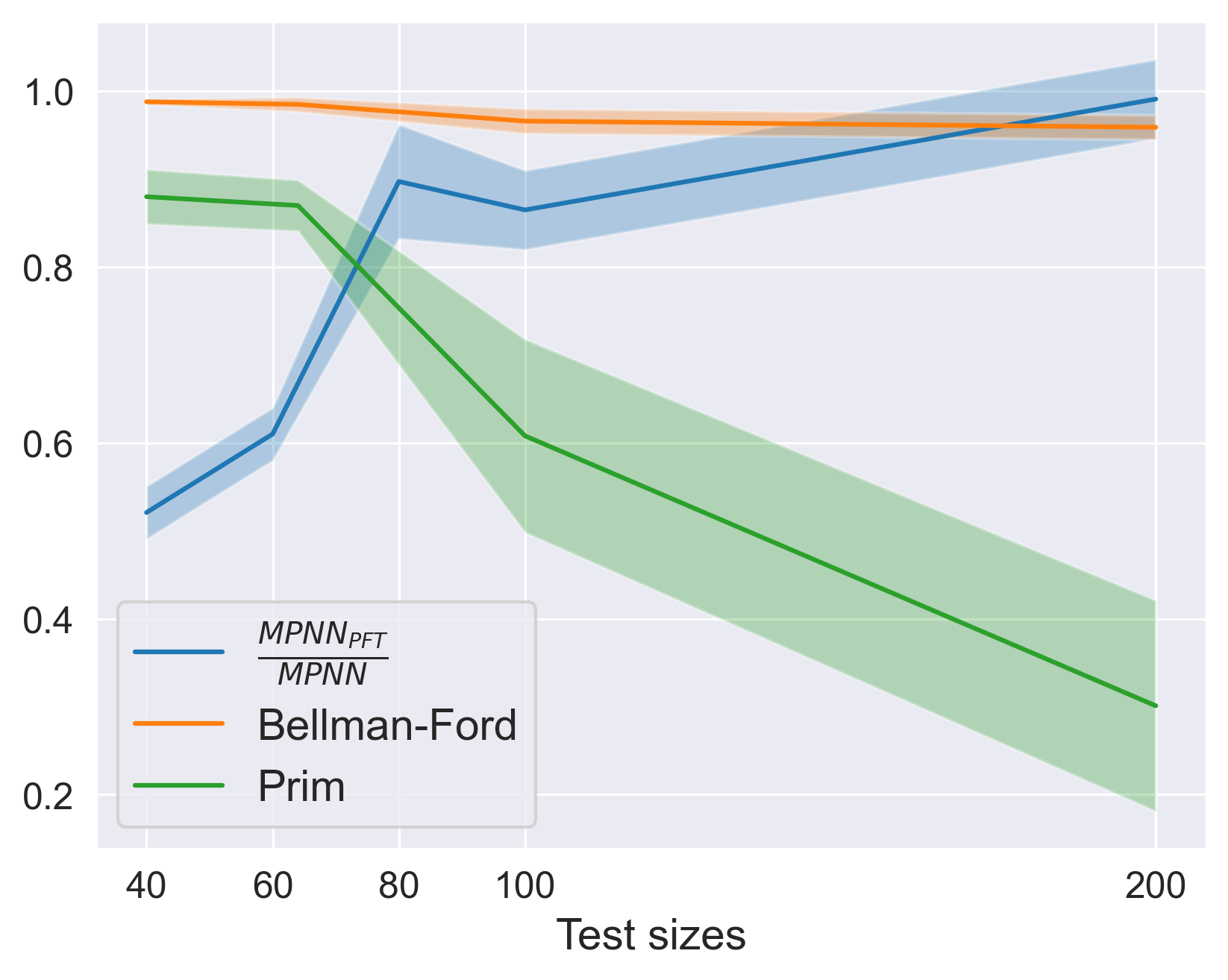}%
    \caption{
        MPNN\textsubscript{PFT} initially outperforms MPNN most likely due to accurate MST predictions (Prim's curve). However, as MPNN\textsubscript{PFT} struggles with larger graphs, the performance gap narrows ($\frac{\text{MPNN\textsubscript{PFT}}}{\text{MPNN}}$ approaches 1).
    }
    \label{fig:MPNNPFTvsMPNN}
  \end{subfigure}\hfill%
  \begin{subfigure}[t]{0.49\textwidth}
    \centering
    \includegraphics[width=0.99\linewidth]{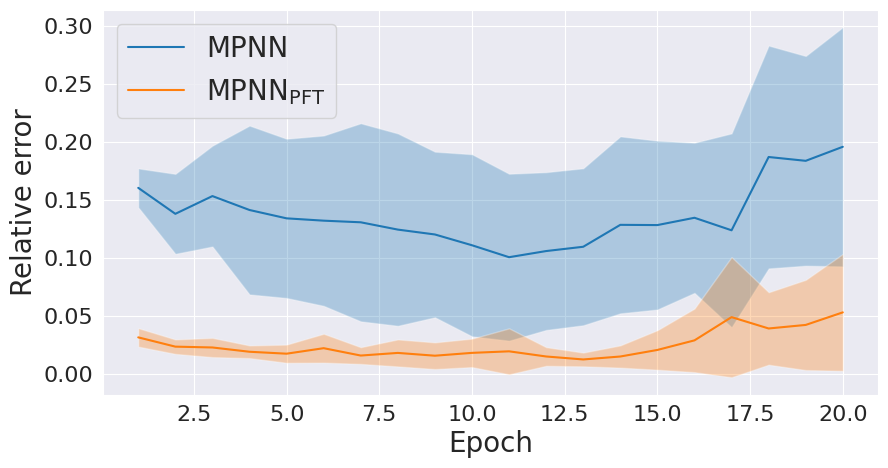}%
    \caption{
        Pre-training on algorithmic reasoning converges faster. Both models use
        the same architecture and hyperparameters.
    }
    \label{fig:PFTFTW}
  \end{subfigure}
  \caption{Comparative analysis: \textbf{(a)} MPNN\textsubscript{PFT} vs. MPNN performance, and \textbf{(b)} Convergence speed with algorithmic reasoning pre-training.}
\end{figure}

\paragraph{Standard transfer does not work} Classical approaches, e.g. as in XLVIN
\citep{deac2021neural}, do not produce good solutions. 
Contrasted to the
fine-tuned model, which usually tends to achieve the best in-distribution
validation loss around epoch 13-14, having one frozen processor tends to
produce only worse performance with training, as evident in \autoref{fig:PFTvsPF}. While for some target
tasks $\mathcal{T}$ it may be desirable to be as faithful as possible to the base
$\mathcal{B}$, our experiments suggest this is not the case here.

\begin{table}[t]
  \footnotesize
  \caption{
      TSP extrapolation relative error across different sizes. All
      models are trained on the same data of graphs up to size 20. \textbf{MPNN}
      denotes our base model without algorithmic knowledge. We added
      a subscript if we \textbf{u}sed \textbf{e}dge \textbf{h}idden state or
      attempted knowledge transfer (cf. \autoref{subsec:transfer}).
      Note, that despite being a heuristic, LKH almost always produced optimal solutions.
  }
  \label{tab:synthetic-results}
  \centering
  \begin{tabular}{lrrrrrr}
    \toprule
    & \multicolumn{6}{c}{\bf Test size}\\
    \cmidrule(r){2-7}
     {\bf Model} & \multicolumn{1}{c}{\it 40} & \multicolumn{1}{c}{\it 60}
    & \multicolumn{1}{c}{\it 80} & \multicolumn{1}{c}{\it 100} & \multicolumn{1}{c}{\it 200} & \multicolumn{1}{c}{\it 1000}\\
    \midrule
     \multicolumn{7}{c}{\bf Beam search with width w=128}\\
    \midrule
     MPNN & $17.7_{\pm 5}\%$ & $23.9_{\pm 3}\%$ & $25.7_{\pm 8}\%$ & $31.9_{\pm 6}\%$ & $38.9_{\pm 7}\%$ & $39.7_{\pm 7}\%$\\
     GPS & $13.9_{\pm 1}\%$ & $26.3_{\pm 2}\%$ & $24.1_{\pm 4}\%$ & $29.7_{\pm 5}\%$ & $35.8_{\pm 4}\%$ & $39.9_{\pm 2}\%$\\
     MPNN\textsubscript{PFT} & $9.1_{\pm 0.1}\%$ & $15.5_{\pm 4}\%$ & $23.1_{\pm 3}\%$ & $28.9_{\pm 2}\%$ & $35.4_{\pm 2}\%$ & $44.5_{\pm 9}\%$\\
     MPNN\textsubscript{PFT+ueh} & $15.4_{\pm 5}\%$ & $23.5_{\pm 8}\%$ & $29.4_{\pm 7}\%$ & $35.7_{\pm 6}\%$ & $37.9_{\pm 3}\%$ & $48.8_{\pm 14}\%$\\
     MPNN\textsubscript{2PROC} & $12.2_{\pm 5}\%$ & $22.1_{\pm 4}\%$ & $28.4_{\pm 6}\%$ & $34.0_{\pm 5}\%$ & $36.6_{\pm 6}\%$ & $38.8_{\pm 3}\%$\\
     MPNN\textsubscript{MTL} & $18.1_{\pm 6}\%$ & $26.2_{\pm 4}\%$ & $31.2_{\pm 5}\%$ & $34.8_{\pm 4}\%$ & $37.1_{\pm 3}\%$ & $38.5_{\pm 3}\%$\\
     \begin{tabular}{@{}l@{}}\citeauthor{joshi2022learning}\\{\scriptsize(AR)}\end{tabular} & $6.2_{\pm 4}\%$ & $37.1_{\pm 12}\%$ & $74.8_{\pm 13}\%$ & $102_{\pm 10}\%$ & $195_{\pm 20}\%$ & $419_{\pm 20}\%$\\
     \begin{tabular}{@{}l@{}}\citeauthor{joshi2022learning}\\{\scriptsize(nAR)}\footnotemark\end{tabular} & $20.6_{\pm 10}\%$ & $62.5_{\pm 11}\%$ & $110_{\pm 17}\%$ & $156_{\pm 24}\%$ & $273_{\pm24}\%$ & $497_{\pm 7}\%$\\
     \midrule
     \multicolumn{7}{c}{\bf Beam search with width w=1280}\\
     \midrule
     MPNN & $14.4_{\pm 4}\%$ & $21.8_{\pm 5}\%$ & $22.4_{\pm 4}\%$ & $31.1_{\pm 6}\%$ & $33.6_{\pm 3}\%$ & $41.2_{\pm 8}\%$\\
     GPS & $11.0_{\pm 1}\%$ & $21.0_{\pm 1}\%$ & $19.0_{\pm 3}\%$ & $23.8_{\pm 5}\%$ & $35.3_{\pm 6}\%$ & $40.1_{\pm 2}\%$\\
     MPNN\textsubscript{PFT} & $7.5_{\pm 1}\%$ & $13.3_{\pm 3}\%$ & $20.1_{\pm 5}\%$ & $26.9_{\pm 3}\%$ & $33.8_{\pm 2}\%$ & $40.5_{\pm 7}\%$\\
     MPNN\textsubscript{PFT+ueh} & $12.9_{\pm 5}\%$ & $20.6_{\pm 6}\%$ & $26.0_{\pm 3}\%$ & $32.0_{\pm 5}\%$ & $37.2_{\pm 3}\%$ &
     $44.3_{\pm 9}\%$\\
     MPNN\textsubscript{2PROC} & $9.5_{\pm 6}\%$ & $18.0_{\pm 5}\%$ & $24.1_{\pm 6}\%$ & $28.8_{\pm 5}\%$ & $36.6_{\pm 6}\%$ & $38.5_{\pm 3}\%$\\
     MPNN\textsubscript{MTL} & $15.5_{\pm 5}\%$ & $26.5_{\pm 6}\%$ & $30.2_{\pm 6}\%$ & $33.4_{\pm 4}\%$ & $36.8_{\pm 5}\%$ & $42.3_{\pm 5}\%$\\
     \begin{tabular}{@{}l@{}}\citeauthor{joshi2022learning}\\{\scriptsize(AR)}\end{tabular} & $3.5_{\pm 3}\%$ &  $33.0_{\pm 11}\%$ & $63.8_{\pm 12}\%$ & $97.6_{\pm 12}\%$ & $193.3_{\pm 17}\%$ & $417_{\pm 4}\%$\\
     \begin{tabular}{@{}l@{}}\citeauthor{joshi2022learning}\\{\scriptsize(nAR)}\footnotemark[4]\end{tabular}& $14.2_{\pm 8}\%$ & $54.7_{\pm 15}\%$ & $100.0_{\pm 15}\%$ & $141.9_{\pm 25}\%$ & $265.5_{\pm 27}\%$ & $500.7_{\pm 23}\%$\\
    \midrule
     \multicolumn{7}{c}{\bf Models not using beam search}\\
    \midrule
     \begin{tabular}{@{}l@{}}\citeauthor{joshi2022learning}\\({\scriptsize AR+greedy})\end{tabular} & $22.1_{\pm 10}\%$ & $57.7_{\pm 13}\%$ & $94.0_{\pm 14}\%$ & $124.0_{\pm 12}\%$ & $219.6_{\pm 22}\%$ & $469_{\pm 17}\%$\\
     \begin{tabular}{@{}l@{}}\citeauthor{joshi2022learning}\\({\scriptsize AR+sampling})\end{tabular} & $9.5_{\pm 5}\%$ & $46.7_{\pm 9}\%$ & $93.1_{\pm 10}\%$ & $137.0_{\pm 14}\%$ & $313.2_{\pm 15}\%$ & $1102_{\pm 1}\%$\\
    \midrule
     \multicolumn{7}{c}{\bf Deterministic baselines}\\
    \midrule
     {\it Greedy} & $31.9_{\pm 12}\%$ & $32.8_{\pm 10}\%$ & $33.3_{\pm 9}\%$ & $30.0_{\pm 6}\%$ & $32.1_{\pm 6}\%$ & $28.8_{\pm 3}\%$\\
     {\it \begin{tabular}{@{}l@{}}Beam search\\{\scriptsize (w=1280)}\end{tabular}} & $19.7_{\pm 8}\%$ & $23.1_{\pm 7}\%$ & $29.4_{\pm 7}\%$ & $29.7_{\pm 5}\%$ & $33.2_{\pm 4}\%$ & $38.9_{\pm 2}\%$\\
     {\it Christofides} & $10.1_{\pm 3}\%$ & $11.0_{\pm 2}\%$ & $11.3_{\pm 2}\%$ & $12.1_{\pm 2}\%$ & $12.2_{\pm 1}\%$ & $12.2_{\pm 0.1}\%$\\
     {\it LKH} & $0.0_{\pm 0.0}\%$ & $0.0_{\pm 0.0}\%$ & $0.0_{\pm 0.0}\%$ & $0.0_{\pm 0.0}\%$ & $0.0_{\pm 0.0}\%$ & $0.01_{\pm 0.0}\%$\\
    \bottomrule
  \end{tabular}
\end{table}

\paragraph{Transfer vs no-transfer} Our main results are presented in
\autoref{tab:synthetic-results}. We start by noting that except on the smallest
test sizes, our model outperformed the architecture
of \citeauthor{joshi2022learning}, often by a significant margin. From our results, the latter architecture tends to fit well the training data distribution but exhibits worse OOD generalisation, as confirmed by results on $\geq$ 60 nodes.

Our next observation is that in almost all cases, pre-training our model to
perform algorithms gives better performance even when compared to a more sophisticated baseline utilising the GPS graph transformer convolution \citep{rampavsek2022recipe}. Even though the performance gap diminishes as the test size increases, \autoref{fig:MPNNPFTvsMPNN} suggests that this trend is influenced by the progressive deterioration of algorithmic reasoning performance as test graph size increases. In particular, one can note that MPNN and MPNN\textsubscript{PFT} perform very similarly for graphs with $\geq$200 nodes. This observation aligns with the point where MPNN\textsubscript{PFT} demonstrates its lowest performance on Prim's algorithm, i.e. $\sim 30\%$. This suggests that building neural algorithmic reasoners that can strongly generalise even for much larger graphs becomes critical to solving CO problems using our approach.

Surprisingly, training two processors, i.e. MPNN\textsubscript{2PROC}, did not yield performance gain when compared to MPNN\textsubscript{PFT}, while outperforming the algorithm-agnostic baseline. In the original formulation from \citeauthor{xhonneux2021transfer} the latent representations of the two processors are summed together. However, summing the two representations was unstable and led to crashes in our experiments. In our experiments, we used mean instead. MPNN\textsubscript{MTL} also exhibited inferior performance compared to other models. In general, our results suggest that transforming knowledge from P to NP is not trivial. Specifically, we note that our pre-trained reasoners emit representations from which we can easily decode steps of Prim and Bellman-Ford. However, transforming this information into a ``good heuristic'' for the TSP can not be achieved by trivially linearly projecting it (MPNN\textsubscript{PF}, see \autoref{fig:PFTvsPF}), or by averaging it with a learnt representation for the TSP (MPNN\textsubscript{2PROC}). Clearly, learning a representation that has to both encode information for P and NP (MPNN\textsubscript{MTL}) did not lead to meaningful representations either, since it performs worse than a simple MPNN. We conclude that the best way of transforming such information to a good performing heuristic for an NP problem is to fine-tune the representations (MPNN\textsubscript{PFT}).

Additionally, we noticed that when compared to a model with an
identical architecture but without algorithmic knowledge, inducing algorithmic
knowledge led to faster convergence and exhibited delayed overfitting compared to its corresponding non-pretrained model, as depicted in Figure \ref{fig:PFTFTW}.

Lastly, we note that we can outperform simpler non-parametric baselines on nearly all test sizes, i.e. {\it greedy} and {\it beam search} and we perform better and comparable to Christofides for 2x and 3x larger graphs. Unfortunately, \emph{all} neural models still fall short at larger graphs and we are outperformed by LKH, which almost always produces optimal solutions.
We hypothesise that with stronger algorithmic generalisation OOD we could match Christofides algorithm and that future studies could investigate integrating LKH with NAR (or vice versa).

\footnotetext{
    To disambiguate non-autoregressive decoding, as in \citet{joshi2022learning}
    from neural algorithmic reasoning we will use nAR and NAR respectively
}

\paragraph{Does it matter what we put in it?} An ``alternative hypothesis''
that one can make is that the improvements are due to the network being trained
on the same data distribution and the algorithms themselves do not matter and
any algorithm would suffice. We present two counterarguments:
\textbf{First}, results in \autoref{app:algosel_wtf} show that selecting
unrelated algorithms is subpar to selecting related ones and may also result
in final bad local minima with poorer generalisation. However, when one is
uncertain about what algorithms to pick, a generalist-like pre-training
\citep{ibarz2022generalist} may still be a viable option. \textbf{Second},
results in \autoref{app:euc-vs-noneuc} show that changing the source data
distribution does not lead to results worse than no transfer and may even
produce better results at larger test sizes. Those two ablations 
suggest it is the algorithmic bias that is useful for learning CO problems.

\subsection{VKC}

\paragraph{Benchmarks} Given the strong performance of the PFT transfer
learning variant and for computational efficiency, we concentrated on comparing
best transfer versus no transfer, foregoing retesting of other transfer options
with VKC. Additionally, we chose two deterministic baselines: the Gon
algorithm, which influenced our algorithm selection, and the critical
dominating set (CDS) heuristic \citep{garciadiaz2017when}. Despite its poorer
approximation ratio, CDS consistently provides good solutions in practice.

\begin{table}[t]
  \footnotesize
  \caption{
      VKC relative error. Each letter in the MPNNs' subscript denotes an algorithm
      pre-trained on: \textcolor{Purple}{F}loyd-Warshall, \textcolor{Purple}{M}inumum,
      \textcolor{Purple}{I}nsertion sort, \textcolor{Purple}{T}ask scheduling, \textcolor{Purple}{B}ellman-Ford,
      \textcolor{Purple}{A}ctivity selection.
  }
  \label{tab:vkc-results}
  \centering
  \begin{tabular}{lrrrrr}
    \toprule
    & \multicolumn{5}{c}{\bf Test size}\\
    \cmidrule(r){2-6}
     {\bf Model} & \multicolumn{1}{c}{\it 40} & \multicolumn{1}{c}{\it 60}
    & \multicolumn{1}{c}{\it 80} & \multicolumn{1}{c}{\it 100} & \multicolumn{1}{c}{\it 200}\\
     \midrule
     MPNN & $15.88_{\pm 1.11}\%$ & $20.85_{\pm 2.64}\%$ & $24.78_{\pm 7.17}\%$ & $26.48_{\pm 6.65}\%$ & $24.63_{\pm 6.50}\%$ \\
     MPNN\textsubscript{FMITB} & $14.73_{\pm 1.21}\%$ & $21.97_{\pm 4.87}\%$ & $23.91_{\pm 4.52}\%$ & $28.50_{\pm 5.06}\%$ & $27.04_{\pm 4.22}\%$ \\
     MPNN\textsubscript{MTAB} & $13.58_{\pm 0.60}\%$ & $16.20_{\pm 3.02}\%$ & $23.10_{\pm 3.99}\%$ & $26.40_{\pm 6.62}\%$ & $26.43_{\pm 3.84}\%$ \\
     \midrule
     \multicolumn{6}{c}{\bf Deterministic baselines}\\
     \midrule
     Farthest First & $41.04_{\pm 14.21}\%$ & $43.89_{\pm 10.70}\%$ & $38.31_{\pm 9.22}\%$ & $36.32_{\pm 7.28}\%$ & $37.91_{\pm 7.91}\%$\\
     CDS & $7.15_{\pm 5.42}\%$ & $7.82_{\pm 4.78}\%$ & $6.49_{\pm 3.65}\%$ & $6.55_{\pm 3.29}\%$ & $5.98_{\pm 2.00}\%$\\
    \bottomrule
  \end{tabular}
\end{table}

\paragraph{Results} \autoref{tab:vkc-results} summarises our results with VKC.
When picking algorithms, solely based on relevance to the task, we obtain
results comparable, or worse than versions without pre-training. Removing the
two poor performing algorithms (Floyd-Warshall, mean accuracy of $\approx23\%$;
Insertion Sort, $\approx43\%$), in favour of an additional greedy (activity
selection, $\approx96\%$) results in a better model, substantially
outperforming the baseline neural model at $2\times$ and $3\times$
extrapolation and tying for the other sizes. This further supports our
hypothesis that generalising performance on algorithms is related to the
performance on the downstream CO task.

Compared to the deterministic baselines, all neural models outperform the
simpler heuristic. Unfortunately, even the best-performing models are not
comparable to the CDS heuristic. We believe this is due to the fact that CDS
consists of multiple, interlaced subroutines (binary search, graph pruning,
etc.) and perhaps a future more modular NAR approach would give further performance boost.

\section{Limitations and future work}

Despite the promising results achieved in our study, there are still areas for
improvement in current algorithmic reasoning networks, including our approach.
In particular, the GNN requires $\mathcal{O}(N)$ message passing steps. This
constraint makes training on very large graphs impractical (\autoref{app:runtime}).
A future work direction we envision is NARs which require
fewer iterations to execute an algorithm. Our intuition is based on
\citet{xu2020what} who, in their experiments of learning to find shortest
paths, noted that a GNN may sometimes achieve satisfactory test accuracy with
fewer iterations than the ground truth algorithm. Such a contribution, however,
is out of the scope of this paper.

\section{Conclusions}

In this study, we explored the transfer of knowledge from algorithms in P to
the travelling salesman problem and vertex k-center, both NP-hard combinatorial
optimisation problems. Our findings revealed that standard transfer learning
techniques did not yield satisfactory results. However, we observed that
certain approaches, which were previously considered to have inferior
performance, actually exhibited superior generalization capabilities and
outperformed models that do not leverage pre-training on algorithms.

In conclusion, our findings provide strong evidence supporting the hypothesis
that incorporating robust algorithmic knowledge is beneficial for achieving
out-of-distribution generalization. By leveraging the insights from algorithms,
we can enhance the performance of reasoning models on complex problems. These
findings open up new avenues for the development of algorithmic reasoners that
can effectively tackle challenging real-world tasks and applications.

\newpage 
\bibliography{ref.bib}
\bibliographystyle{abbrvnat}

\newpage
\begin{appendix}
    \section{Runtime}\label{app:runtime}
    Runtimes have been obtained on a A100 40GB GPU.
    \begin{table}[h]
    \footnotesize
    \caption{Training and inference time on algorithms. This measures the efficiency of the pre-training phase. Time is relative to a single batch. Note that inference graphs are 4$\times$ the size of training graphs}\label{tab:train_inference_times_algos}
    \centering
    \begin{tabular}{lcccc}
        \toprule
        & \multicolumn{2}{c}{\textbf{Training time (s)}} & \multicolumn{2}{c}{\textbf{Inference time (s)}} \\
        & Erdős–Rényi & Cliques & Erdős–Rényi & Cliques \\
        \midrule
        Bellman-Ford & $0.057$ & $0.103$ & $0.237$ & $0.291$ \\
        Dijkstra & $0.160$ & $0.184$ & $1.469$ & $3.004$ \\
        MST-Prim & $0.162$ & $0.186$ & $1.499$ & $2.999$ \\
        MST-Kruskal & $1.576$ & $2.073$ & $32.997$ & $44.957$ \\
        Floyd-Warshall & $0.231$ & OOM (16GB) & $9.827$ & OOM(16GB) \\
        \bottomrule
    \end{tabular}
    \end{table}

    \begin{table}[h]
      \footnotesize
      \caption{Inference time of TSP solutions (in seconds). Time is relative to a single TSP of size $n$.}\label{tab:inference_time}
      \centering
      \begin{tabular}{llccccccc}
        \toprule
        && \multicolumn{5}{c}{\bf Test size}\\
        \cmidrule(r){3-9}
        & {\bf Model} & {\it 20} & {\it 40} & {\it 60}
        & {\it 80} & {\it 100} & {\it 200} & {\it 1000}\\
        \midrule
        \texttt{w=128}
        & MPNN & $0.043$ & $0.090$ & $0.130$ & $0.179$ & $0.242$ & $0.800$ & $65.557$\\
        \texttt{w=1280}
        & MPNN & $0.043$ & $0.089$ & $0.131$ & $0.181$ & $0.225$ & $0.839$ & $67.009$\\
        \midrule
        & {\it Greedy} & $0.001$ & $0.002$ & $0.004$ & $0.009$ & $0.019$ & $0.021$ & $0.047$\\
        & {\it Christofides} & $0.004$ & $0.012$ & $0.411$ & $0.094$ & $0.101$ & $0.714$ & $57.584$\\
        \bottomrule
      \end{tabular}
    \end{table}

\section{Algorithm selection rationale}\label{app:algosel}

\paragraph{Travelling Salesman Problem} TSP is defined on an undirected full
graph, often originating from 2D Euclidean space. We thus considered graph and
geometry algorithms in CLRS-30. Many of the algorithms were excluded as they
were either trivial to solve on a full graph (e.g., BFS/articulation points),
or were defined only on directed (acyclic) graphs (e.g., DFS, topological
sort). We also excluded algorithms with long rollouts: e.g., in CLRS-30,
MST-Kruskal and Dijkstra's algorithms have (much) longer trajectories than
MST-Prim and Bellman-Ford, despite having equal or better theoretical time
complexity. Such long-trajectory algorithms are impractical for our
application, as they result in higher training/inference times
(\autoref{app:runtime}, \autoref{tab:train_inference_times_algos}). The
resulting list, in the end, contained 3 relevant algorithms -- Bellman-Ford,
MST-Prim and Graham's scan. Unfortunately in our initial experiments,
integrating Graham led to poor performance and it was removed from our final
models. We suspect this was due to the fact that NAR architectures do not
incorporate invariance to Euclidean transformations
\citep{bronstein2021geometric}.

\paragraph{Vertex k-center problem} The Vertex k-center problem employs
a straightforward 2-approximation heuristic, commonly known as the Gon
algorithm \citep{gonzalez1985clustering, dyer1985asimple}. This heuristic
involves randomly selecting the first station and then greedily choosing the
vertex that is farthest from any of the currently chosen vertices. To learn
this heuristic, the model needs to perform shortest path calculations between
nodes and understand concepts like maximum finding (or minimum of negatives)
and/or sorting. Considering the greedy algorithm nature of the Gon algorithm,
we decided to pre-train our model on Bellman-Ford, Minimum finding, and two
CLRS-30 greedy problems: activity selection \citep{gavril1972algorithms} and
task scheduling \citep{lawler2001combinatorial}. The reason to choose Bellman-Ford
over Floyd-Warshall (an all-pairs shortest path algorithm) and Minimum over
Insertion sort is that Floyd-Warshall and Insertion sort have much poorer
performance.

\section{On selecting unrelated algorithms}\label{app:algosel_wtf}

\begin{table}[t]

  \footnotesize
  \caption{
      TSP relative error comparison between no-transfer, transferring unrelated
      algorithms and transferring related algorithms. When transferring, the
      experimental setup matches MPNN\textsubscript{PFT}, only algorithms differ
  }
  \label{tab:relvsunrel}
  \centering
  \begin{tabular}{lrrrrrr}
    \toprule
    & \multicolumn{6}{c}{\bf Test size}\\
    \cmidrule(r){2-7}
     {\bf Model} & \multicolumn{1}{c}{\it 40} & \multicolumn{1}{c}{\it 60}
    & \multicolumn{1}{c}{\it 80} & \multicolumn{1}{c}{\it 100} & \multicolumn{1}{c}{\it 200} & \\
    \midrule
     \multicolumn{7}{c}{\bf Beam search with width w=128}\\
    \midrule
     MPNN & $17.7_{\pm 5}\%$ & $23.9_{\pm 3}\%$ & $25.7_{\pm 8}\%$ & $31.9_{\pm 6}\%$ & $38.9_{\pm 7}\%$ & \\
    Unrelated & $12.3_{\pm 2.2}\%$ & $17.5_{\pm 3.1}\%$ & $22.3_{\pm 3.1}\%$ & $26.4_{\pm 2.6}\%$ & $199.0_{\pm 373.3}\%$ & \\
     Related & $9.1_{\pm 0.1}\%$ & $15.5_{\pm 4}\%$ & $23.1_{\pm 3}\%$ & $28.9_{\pm 2}\%$ & $35.4_{\pm 2}\%$ & \\
    \bottomrule
  \end{tabular}
\end{table}

To evaluate the impact of selecting an appropriate algorithm for training, we
conducted the following additional experiment. We utilized our best TSP
algorithm transfer configuration while pre-training on three different
algorithms: \emph{Breadth-First Search} (which is trivial on fully-connected
graphs), \emph{Topological Sorting} (defined for directed acyclic graphs), and
\emph{Longest Common Subsequence length} (a string algorithm). Results are
presented in \autoref{tab:relvsunrel}. While transferring unrelated algorithms
still brings improvements and is comparable at sizes 80 and 100, the model starts to
``explode'' at size 200 and it cannot provide as optimal solutions at sizes 40 and 60.

\section{Transfer from non-euclidean data distribution}\label{app:euc-vs-noneuc}

To evaluate to what extent the data distribution matters when pre-training, we
reran our PFT experiment, but this time pre-training the algorithms on graphs
from the Erdős-Renyi distribution. Results are given in
\autoref{tab:euc-vs-noneuc}.

\begin{table}[ht]
  \footnotesize
  \caption{
      Pre-training on Euclidean data versus pre-training on Erdős-Renyi graphs.
  }
  \label{tab:euc-vs-noneuc}
  \centering
  \begin{tabular}{lrrrrrr}
    \toprule
    & \multicolumn{6}{c}{\bf Test size}\\
    \cmidrule(r){2-7}
     {\bf Model} & \multicolumn{1}{c}{\it 40} & \multicolumn{1}{c}{\it 60}
    & \multicolumn{1}{c}{\it 80} & \multicolumn{1}{c}{\it 100} & \multicolumn{1}{c}{\it 200} & \multicolumn{1}{c}{\it 1000}\\
    \midrule
     \multicolumn{7}{c}{\bf Beam search with width w=128}\\
    \midrule
     MPNN & $17.7_{\pm 5}\%$ & $23.9_{\pm 3}\%$ & $25.7_{\pm 8}\%$ & $31.9_{\pm 6}\%$ & $38.9_{\pm 7}\%$ & $39.7_{\pm 7}\%$\\     MPNN\textsubscript{NON\_EUC} & $10.5_{\pm 3}\%$ & $17.3_{\pm 4}\%$ & $22.6_{\pm 3}\%$ & $25.1_{\pm 3}\%$ & $31.0_{\pm 2}\%$ &
     $36.9_{\pm 2}\%$\\
     MPNN\textsubscript{EUC} & $9.1_{\pm 0.1}\%$ & $15.5_{\pm 4}\%$ & $23.1_{\pm 3}\%$ & $28.9_{\pm 2}\%$ & $35.4_{\pm 2}\%$ & $44.5_{\pm 9}\%$\\
     \midrule
     \multicolumn{7}{c}{\bf Beam search with width w=1280}\\
     \midrule
     MPNN & $14.4_{\pm 4}\%$ & $21.8_{\pm 5}\%$ & $22.4_{\pm 4}\%$ & $31.1_{\pm 6}\%$ & $33.6_{\pm 3}\%$ & $41.2_{\pm 8}\%$\\
     MPNN\textsubscript{NON\_EUC} & $8.5_{\pm 3}\%$ & $16.0_{\pm 3}\%$ & $20.0_{\pm 3}\%$ & $23.0_{\pm 2}\%$ & $30.3_{\pm 2}\%$ & $37.5_{\pm 3}\%$\\
     MPNN\textsubscript{EUC} & $7.5_{\pm 1}\%$ & $13.3_{\pm 3}\%$ & $20.1_{\pm 5}\%$ & $26.9_{\pm 3}\%$ & $33.8_{\pm 2}\%$ & $40.5_{\pm 7}\%$\\
    \bottomrule
  \end{tabular}
\end{table}

\section{Benchmarking versus CP/MIP}\label{app:gurobiowned}
\newcolumntype{L}{>{$}l<{$}}
\begin{table}[ht]
    \centering
    \begin{tabular}{l L L L L L L}
        \toprule
        & \multicolumn{6}{c}{\bf Test size}\\
        \cmidrule(r){2-7}
        \textbf{Model} & 40 & 60 & 80 & 100 & 200 & 1000 \\
        \midrule

        \multicolumn{7}{c}{\bf Relative time limit}\\
        \midrule
        Gurobi ($1\times$) & 102.0_{\pm 133}\% & \bot & \bot & \bot & \bot & \bot \\
        Gurobi ($3\times$) & 15.0_{\pm 19}\% & 35.0_{\pm 46}\% & \bot & \bot & \bot & \bot \\
        Gurobi ($5\times$) & 6.3_{\pm 9}\% & 21.9_{\pm 15}\% & 25.7_{\pm 18}\% & \bot & \bot & \bot \\
        \midrule
        \multicolumn{7}{c}{\bf Absolute time limit}\\
        \midrule
        Gurobi (3s) & 1.0_{\pm 2}\% & 4.1_{\pm 4}\% & 13.1_{\pm 11}\% & 26_{\pm 27}\% & \bot & \bot \\
        Gurobi (5s) & 0.4_{\pm 1}\% & 2.6_{\pm 3}\% & 7.8_{\pm 7}\% & 15_{\pm 15}\% & \bot & \bot \\
        Gurobi (10s) & 0.1_{\pm 0}\% & 0.6_{\pm 1} & 5.8_{\pm 5}\% & 7_{\pm 6}\% & \bot & \bot \\
        Gurobi (30s) & - & - & - & - & 18_{\pm 12}\% & \bot \\
        Gurobi (60s) & - & - & - & - & 12_{\pm 9}\% & \bot \\
        Gurobi ($\infty$s) & - & - & - & - & - & OOM (30GB)\\
        \bottomrule
    \end{tabular}
    \caption{
        Relative error of the solution Gurobi provided, \emph{given the time
        limit}. Relative time limit is respectively $1\times, 3\times$ and
        $5\times$ the runtime of our model, as taken from
        \autoref{tab:inference_time}. $\bot$ means the solver did not produce
        a solution within the time constraint. We decided not to run
        experiments with $-$ -- while we have not seen it, based on previous
        experiments and relative improvements we assume Gurobi would have found
        the optimal solutions.
    }
    \label{tab:mytable}
\end{table}

We will start this appendix by noting that CP and MIP solvers would always
produce the optimal solution \emph{if provided with sufficient compute time}.
We therefore will be comparing with such solvers after setting
a time limit for their computation. As such solvers leverage CPU multithreading,
all experiments will be performed on a 16-thread AMD Ryzen 5000 series CPU. Note, that in vanilla TSP those approaches may also be costly to compute, and CP works better for problems having a lot of combinatorial constraints, such as TSP with time windows \citep{savelsbergh1985local}.

Our initial experiment with more sophisticated baselines and constraint
programming concretely was using Gecode \footnote{https://www.gecode.org/}
interfaced through \texttt{Zython}\footnote{https://github.com/ArtyomKaltovich/zython}.
Unfortunately solving \emph{a single TSP instance of size 40} took nearly 2600
seconds and the dataset consists of 1000 such samples. We therefore did not
continue further experiments and switched to Gurobi, which could solve examples
of this size in just a couple of seconds.

We performed two kinds of experiments with Gurobi. One with a relative time
limit ($1\times, 3\times, 5\times$ of our model inference time) and one with an
absolute time limit (always larger than corresponding $5\times$). Across all
variants, the TSP problem was formulated using the Miller-Tucker-Zemlin integer
programming formulation \citep{miller1960integer} and we used the default
Gurobi configuration (by not setting any environmental variables of
PuLP\footnote{\url{https://github.com/coin-or/pulp/tree/master}}).
Results are presented
in \autoref{tab:mytable}. At $1\times$ and $3\times$ Gurobi produced worse
solutions than our best model and only produced better solutions for the smallest
size at $5\times$. Solution quality did improve as time was increased, but at
the larger sizes, Gurobi needed more than 30s to produce a solution and ran out
of memory (OOM) at the largest size, without producing a solution.

Unfortunately, we still fall short of Concorde, which always found the optimal
solution even at the strictest time constraints, even though it does not always
prove the optimality. Note, however, that Concorde is highly specialised in
solving TSP instances. It is our belief that models aligned to parallel
algorithms \citep{engelmayer2023parallel} would help improve runtimes, leading
to GNNs eventually outperforming Concorde.

\end{appendix}

\end{document}